# PERFORMANCE ANALYSIS OF PID, PD AND FUZZY CONTROLLERS FOR POSITION CONTROL OF 3-DOF ROBOT MANIPULATOR

[1,2]Usman Kabir, [2]Mukhtar Fatihu Hamza [2]Ado Haruna and [3]Gaddafi Sani Shehu
[1]Department of Computer Engineering, Hussaini Adamu Polytechnic, Kazaure
[2]Department of Mechatronics Engineering, Bayero University, Kano
[3]Department of Electrical Engineering, Ahmadu Bello University, Zaria
[1]uk4uk70@gmail.com, [2]aharuna.mct@buk.edu.ng, [3]gsshehu@abu.edu.ng

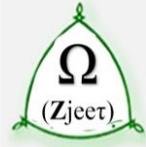
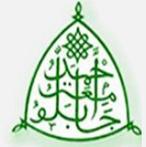

## ABSTRACT

*Robot manipulators are extensively used in industrial applications because of their immense importance such as in constructions automation. Therefore, designing controllers to suit the intensive purpose of the application is one of the major challenges for control researchers. This paper presents comparative analysis of three controllers on a 3-DOF robot manipulator. The proportional integral derivative (PID), proportional derivative (PD) and fuzzy logic (FL) controllers were designed and applied to each of the link of the robot by simulation. The performance of each control method was assessed using the transient and steady state response characteristics. Comparisons of the results obtained, PID and PD performed better in terms of Rise Time and Settling Time while the FLC exhibited reduced overshoot*
.

## 1. INTRODUCTION

Industrial robot manipulators have widespread application such as pick and place, welding, painting, material handling and many more [1]. Due to huge applications of these robots, the need to design and test different control approaches to improve performance through increased precision has become a necessity and an important research area [2]. To achieve a good performance and tracking control design, the physical characteristics or the mathematical equation for the system behaviour has to be developed (i.e dynamic model of the system). This deals with the determination of dynamic equation and mapping forces exerted on the system parts as well as with the motion of the robot manipulator (i.e its joint position, velocity and acceleration). These derived mathematical equations constitute the dynamic model of robot manipulator [3], [4]

Another vital aspect of robot manipulators are the control design and which is an area of interest where various control methodologies have been proposed. The motion controls of the robot manipulator are usually designed to achieve high speed operation, minimum tracking error, disturbance rejection and multi-functionality [5], [6]

Sliding mode control (SMC), sliding mode fuzzy control (SMFC) and adaptive sliding mode fuzzy control (ASMFC) was implemented on robot manipulator and compared. It was found that ASMFC is fairly better than SMC and SMFC in terms of rise time, overshoot and settling time [7]. In [8] demonstrated the application of neural network and fuzzy logic system for on-line identification, where an FLC was used for robot position control and neural network for on-line identification during the control of the system. Modelling and computed torque control of 6-DOF robotic arm was presented in [9].The tracking responses were obtained by applying different inputs to each link. The dynamic modelling and motion control of three link manipulator was presented in [10] where a camera for capturing the motion of the user was employed and PD controller was used for the system control.

Performance comparison of PID and FLC was presented in [11] where difference defuzzification strategies of FLC were applied and all surpassed that of centre of gravity and PID demonstrate the superior performance in terms of settling time, as compared to FLC, while FLC performed better in terms of overshoot but both controllers manage to converge to the desired output.

The conventional PID controller exhibits good performance for linear system and it is widely employed in industry due to its simple structure and robustness in different operation conditions. However, the accurate tuning of the parameters of PID becomes difficult because most of industrial plants are highly complex and have some issues such as nonlinearities, time delays, and higher orders. Due to the complexity of most industrial plants and the limitation of PID controller, an unprecedented interest was diverted to the applications of the FLC. This is because it uses expert knowledge and its control action is described by linguistic rules. Also, the FLC does not require the complete mathematical model of the system to be controlled and it can work properly with nonlinearities and uncertainties. In this work, a comparative analysis of PID, PD and FLC on a 3-DOF planar robot manipulator is presented. The three controllers were designed and implemented on the system which constitutes the shoulder, elbow and wrist for trajectory tracking. Performance comparisons of the controllers were carried out on each link by evaluating the transient and steady state characteristics.

## 2. DYNAMICS MODEL

Dynamic Analysis of a robot manipulator is an important stage to determine the relationship between joint torques/force applied by actuators and the position, velocity and acceleration of the system with respect to time which describe the dynamic parameters in order to efficiently design, control and simulate the system [1], [12]. The dynamic equations of the system are resolve from Figure (1) Using Lagrange approach and is usually presented the final output as

$$\tau = M(q)\ddot{q} + C(q\dot{q})\dot{q} + G(q) \ldots \quad (1)$$

Where, $\tau$ represent the control input torque, $M(q)\ddot{q}$ represent $n \times n$ symmetric and positive definite inertia matrix, $C(q\dot{q})\dot{q}$ represent $n \times 1$ vector of centrifugal and coriolis and $G(q)$ represent $n \times 1$ vector of gravitational torque.

q is n × 1 vector of joint position, $\dot{q}$ – is a n × 1 of joint velocity, and $\ddot{q}$ – is × 1 of joint acceleration

$n$ =Represent the number of degree of freedom of the robot manipulator [1], [8], [10], [13],

The final dynamic equations which were obtained through mathematical derivation of the system in Figure 1 were presented based on the matrix notation of equation (1)

$$M(q) = \begin{bmatrix} a_{11} & a_{12} & a_{13} \\ a_{21} & a_{22} & a_{23} \\ a_{31} & a_{32} & a_{33} \end{bmatrix}, \quad C(q,\dot{q}) = \begin{bmatrix} b_1 \\ b_2 \\ b_3 \end{bmatrix}, \quad g(q) = \begin{bmatrix} g_1 \\ g_2 \\ g_3 \end{bmatrix} \ldots \quad (2)$$

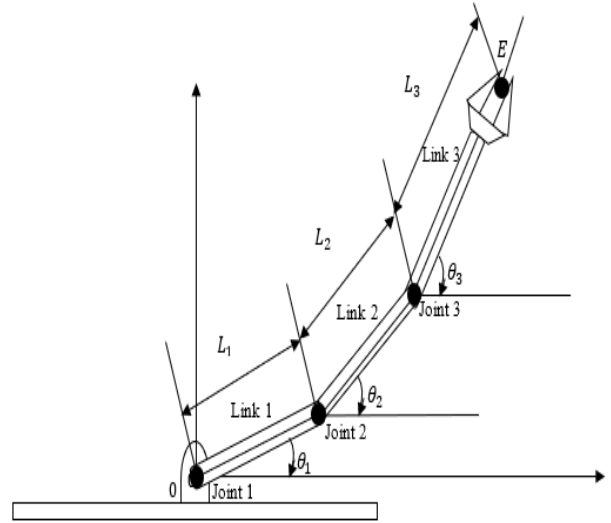

Figure 1: Schematic Diagram of 3-DOF Robot Manipulator

where,

$a_{11} = \{m_1 L_1^2 + m_2(L_1^2 + 2L_1L_2 cos\theta_2 + L_2^2) + m_3(L_1^2 + 2L_1L_2 cos\theta_2 + L_1L_3 cos(\theta_2 + \theta_3) + L_2^2 + 4L_2L_3 cos\theta_3 + 2L_3^2) + J_1 + J_2 + J_3\} \ldots \quad (3)$

$a_{12} = +\{m_2(L_1L_2 cos\theta_2 + L_2^2) + m_3(2L_1L_2 cos\theta_2 + L_1L_3 cos(\theta_2 + \theta_3) + L_2^2 + 4L_2L_3 cos\theta_3 + 2L_3^2) + J_2 + J_3\} \ldots \quad (4)$

$a_{13} = \{m_3(L_1L_3 cos(\theta_2 + \theta_3) + 2L_2L_3 cos\theta_3 + 2L_3^2)\} \ldots \quad (5)$

$a_{21} = \{m_2(L_1L_2 cos\theta_2 + L_2^2) + m_3(2L_1L_2 cos\theta_2 + L_1L_3 cos(\theta_2 + \theta_3) + L_2^2 + 4L_2L_3 cos\theta_3 + 2L_3^2) + J_2 + J_3\} \ldots \quad (6)$

$a_{22} = \{m_2 L_2^2 + m_3(L_2^2 + 4L_2L_3 cos\theta_3 + 2L_3^2) + J_2 + J_3\} \ldots \quad (7)$

$a_{23} = \{m_3(2L_2L_3 cos\theta_3 + 2L_3^2) + J_3\} \ldots \quad (8)$

$a_{31} = \{m_3(L_1L_3 cos(\theta_2 + \theta_3) + 2L_2L_3 cos\theta_3 + 2L_3^2)\} \ldots \quad (9)$

$a_{32} = \{m_3(2L_2L_3 cos\theta_3 + 2L_3^2) + J_3\} \ldots \quad (10)$

$a_{33} = \{2m_3 L_3^2 + J_3\} \ldots \quad (11)$

$b_1 = -m_2 L_1 L_2 \ (2\dot{\theta}_1\dot{\theta}_2 + \dot{\theta}_2^2)\sin \theta_2 - m_3 L_1 L_2 \ (2\dot{\theta}_1\dot{\theta}_2 + \dot{\theta}_2^2)\sin \theta_2 - m_3 L_1 L_3 \ (2\dot{\theta}_1\dot{\theta}_2 + \dot{\theta}_2^2 + 2\dot{\theta}_2\dot{\theta}_3 + 2\dot{\theta}_1\dot{\theta}_3 + \dot{\theta}_3^2) \ sin(\theta_2 + \theta_3) - 2L_2 L_3 \ (2\dot{\theta}_1\dot{\theta}_3 + 2\dot{\theta}_2\dot{\theta}_3 + \dot{\theta}_3^2)\sin\theta_3 \ldots \quad (12)$

$b_2 = -m_2 L_1 L_2 \ (\dot{\theta}_1\dot{\theta}_2)\sin \theta_2 - m_3 L_1 L_2 \ (\dot{\theta}_1\dot{\theta}_2)\sin \theta_2 - m_3 L_1 L_3 \ (\dot{\theta}_1\dot{\theta}_2 + \dot{\theta}_2\dot{\theta}_3)sin(\theta_2 + \theta_3) - 2L_2 L_3 \ (2\dot{\theta}_1\dot{\theta}_3 + 2\dot{\theta}_2\dot{\theta}_3 + \dot{\theta}_3^2)\sin\theta_3 + m_2 L_1 L_2 \ (\dot{\theta}_1^2 + \dot{\theta}_1\dot{\theta}_2)\sin \theta_2 + m_3 L_1 L_2 \ (\dot{\theta}_1^2 + \dot{\theta}_1\dot{\theta}_2)\sin \theta_2 + m_3 L_1 L_3 \left(\dot{\theta}_1^2 + \dot{\theta}_1\dot{\theta}_2 + \dot{\theta}_1\dot{\theta}_3\right) sin(\theta_2 + \theta_3) \ldots \quad (13)$

$b_3 = -m_3 L_1 L_3 \ (\dot\theta_1 \dot\theta_2 + \dot\theta_1 \dot\theta_3) sin(\theta_2 + \theta_3) - 2m_3 L_2 L_3 \ (\dot\theta_1 \dot\theta_3 + \dot\theta_2 \dot\theta_3) sin\theta_3 + m_3 L_1 L_3 \left(\dot\theta_1^2 + \dot\theta_1 \dot\theta_2 + \dot\theta_1 \dot\theta_3\right) sin(\theta_2 + \theta_3) + 2m_3 L_2 L_3 \ (\dot\theta_1^2 + 2\dot\theta_1 \dot\theta_2 + \dot\theta_1 \dot\theta_3 + \dot\theta_2^2 + \dot\theta_2 \dot\theta_3) \ sin\theta_3 \ldots$  (14)

$g_1 = m_1 L_1 g cos\theta_1 + m_2 g(L_1 cos\theta_1 + L_2 cos(\theta_1 + \theta_2)) + m_3 g(L_1 cos\theta_1 + L_2 cos(\theta_1 + \theta_2) + L_3 cos(\theta_1 + \theta_2 + \theta_3)) \ldots$  (15)

$g_2 = m_2 g(L_2 cos(\theta_1 + \theta_2)) + m_3 g(L_2 cos(\theta_1 + \theta_2) + L_3 cos(\theta_1 + \theta_2 + \theta_3)) \ldots$  (16)

$g_3 = m_3 L_3 g \ cos(\theta_1 + \theta_2 + \theta_3) \ldots$  (17)

The block diagram was designed and simulated using Matlab simulink environment and it is presented in Figure 2. Also Open loop response for all the three links are presented in figure 3.

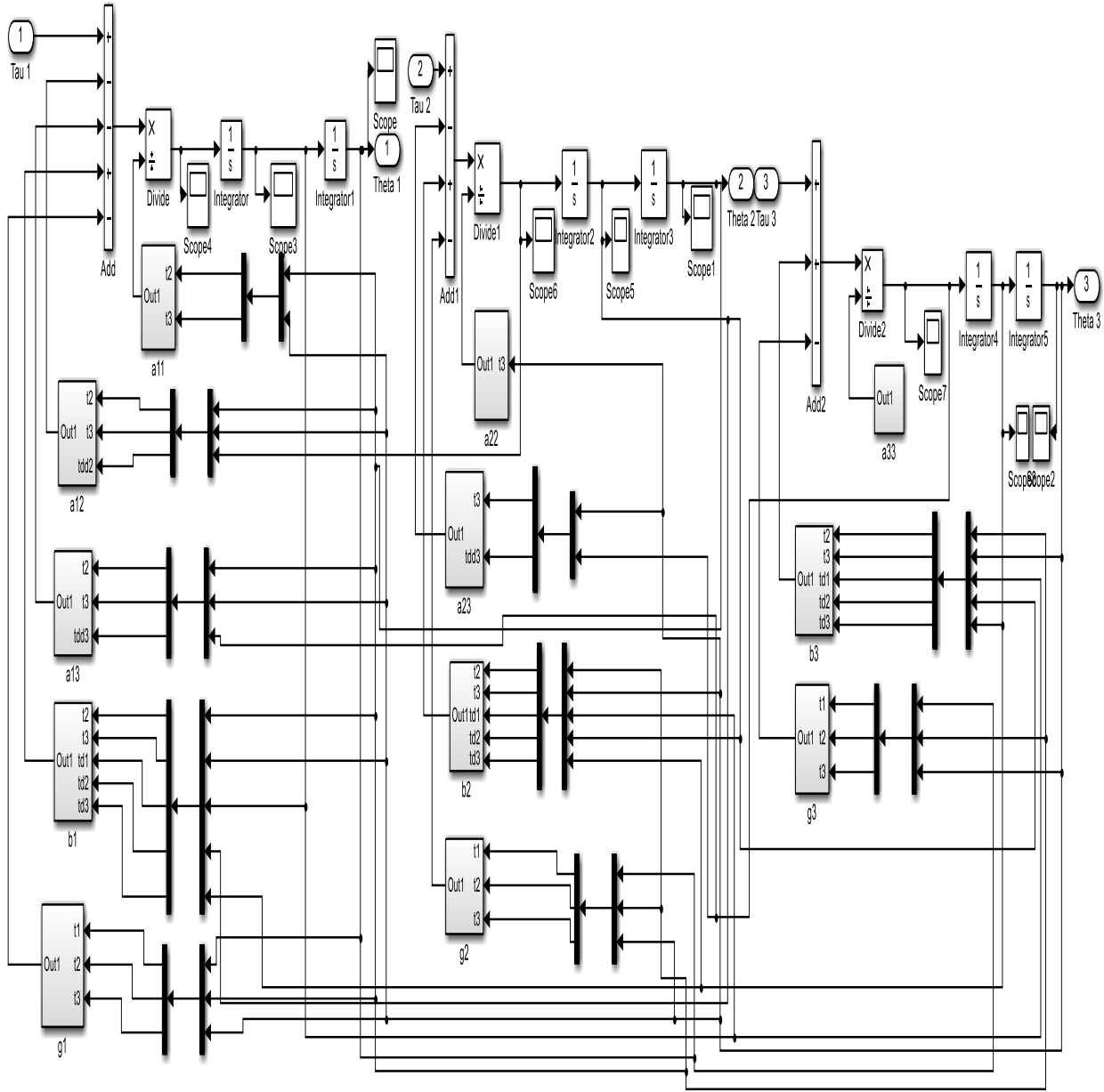

Figure 2: Matlab Model of 3-DOF Robot Manipulator

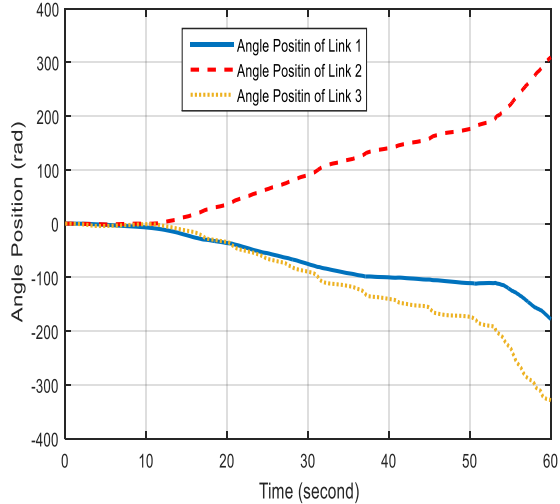

Figure 3: Open Loop Response for the Three Links

## 3. PD AND PID CONTROLLER DESIGN

In this section, PD and PID controllers are implemented for adequate control of the 3-DOF Robot manipulator. Three PD and PID controllers were designed one for each link because all the three links depend on each other.

The PD and PID controllers have the following form respectively

$$u_i = k_{p_i} e_i + k_{d_i} \dot{e}_i \ldots \quad (18)$$

$$u_i = k_{p_i} e_i + k_{I_i} \ddot{e}_i + k_{d_i} \dot{e}_i \ldots \quad (19)$$

Where, $k_p$ is proportional gain, $k_I$ is integral gain, $k_d$ is derivative gain,

$e_i$ is angle error obtained from $\theta_{r_i} - \theta_i$,

$\dot{e}_{d_i}$ is derivative error obtained from $\theta_{r_i} - \theta_{d_i}$,

$\ddot{e}_{I_i}$ is integral error obtained from $\theta_{r_i} - \theta_{I_i}$ and $e_{r_i}$ is reference input

In case of PD and PID $k_{p_i}, k_{I_i}, k_{d_i}$ are adjusted to have a satisfactory result

## 4. FUZZY LOGIC CONTROLLER DESIGN

The operational techniques for designing fuzzy logic controller (FLC) constitute four main units as shown in Figure 4 such units are Fuzzification, inference engine, knowledge base and deffuzification. The fuzzification makes the physical/crisp input data compatible with fuzzy control rule base in core of the controller, inference engine perform the control actions in fuzzy terms according to the information provided by the fuzzification, knowledge base comprise rule base and data base and defuzzification is the inverse of fuzzification [5], [11], [14], [15].

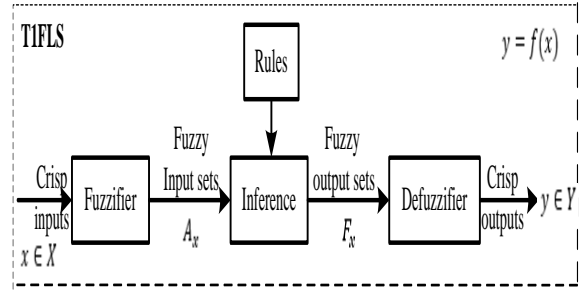

Figure 4: Constituents of Fuzzy Logic Controller

In this work, the Sugeno method was adopted and two inputs were used, these inputs are control input for position error and derivative of position error of the links and one linear output.

The fuzzy controller rule base are tabulated in Table (1) and 9 rule base were formulated as presented below

Table 1: Rules Notation

| $\dot{e}/e$ | P | Z | N |
|---|---|---|---|
| P | PB | P | Z |
| Z | P | Z | N |
| N | Z | N | NB |

where, P is positive, N is negative, Z is zero, PB is positive big, and NB is negative big

### 4.1 Controller Rules Base

IF THEN Rule base are generated base of the observations of the system behaviour. Table 1 described the choices of the linguistics variables eg P, Z, and N which lead to the generation of 9 rules base as shown below. From table 1 it can be seen that "IF the position error and derivative of position error are positive THEN a Big Positive (PB) force is required. Three membership set for the two inputs and five for the output are used to describe all the linguistic variables as shown in Figure 5 (a), (b) and (c) respectively

1) If $e$ is P and $\dot{e}$ is P then $u$ is PB
2) If $e$ is P and $\dot{e}$ is Z then $u$ is P
3) If $e$ is P and $\dot{e}$ is N then $u$ is Z
4) If $e$ is Z and $\dot{e}$ is P then $u$ is P
5) If $e$ is Z and $\dot{e}$ is Z then $u$ is Z
6) If $e$ is Z and $\dot{e}$ is N then $u$ is N
7) If $e$ is N and $\dot{e}$ is P then $u$ is Z

8) If $e$ is N and $\dot{e}$ is Z then $u$ is N
9) If $e$ is N and $\dot{e}$ is N then $u$ is NB

Figure 5 (a), (b) and (c) shows the membership function for the inputs and the output from the Matlab Simulink

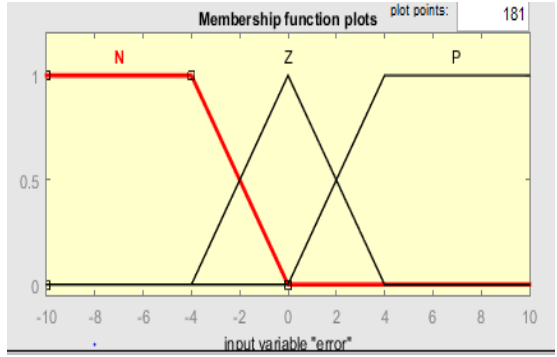

(a)

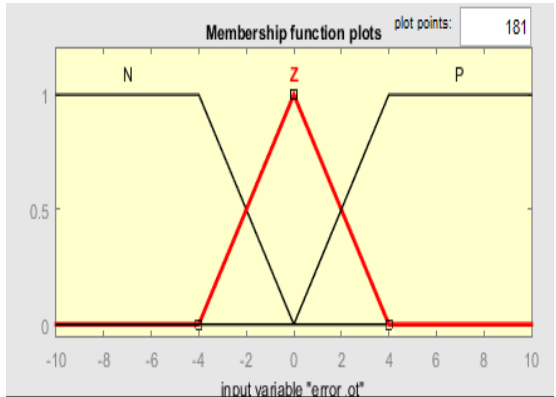

(b)

(c)

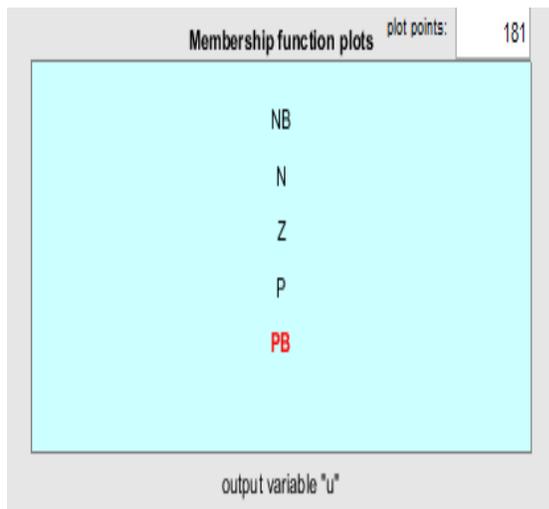

Figure 5: Membership Function (a) Input Error($e$), (b) Input Error dot ($\dot{e}$) and (c) Output ($u$)

### 4.2 Positions Error Set-Point Tracking

The position error is defined as $e_i = \theta_r - \theta_c$ where, $\theta_r$ is reference input (set-point), k is gain and $\theta_c$ is controller system output as illustrate in Figure (6)

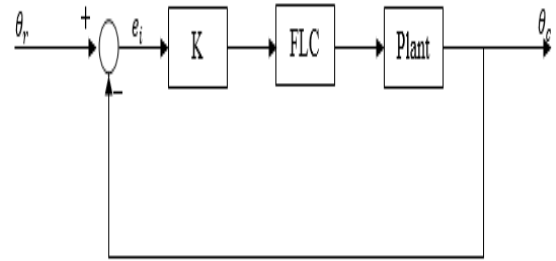

Figure 6: Closed Loop Set Point Tracking

### 4. SIMULATION RESULTS AND SYSTEM PARAMETERS

In this section the system parameters were presented, PD, PID and Sugeno Fuzzy controllers are also implemented in Matlab/Simulink environment

### 4.2 System Parameters

Table 2: System Parameters

| Variables | Values |
|---|---|
| $m_1$ | 1kg |
| $m_2$ | 1kg |
| $m_3$ | 1kg |
| $L_1$ | 0.5m |
| $L_2$ | 0.5m |
| $L_3$ | 0.5m |
| $J_1$ | 0.5kgm² |
| $J_2$ | 0.5kgm² |
| $J_3$ | 0.5kgm² |
| g | 9.81m/s² |

Table 2 presented the system parameters used, where $m_1$, $m_2$ and $m_3$ represent mass of Link1, Link2 and Link3 of the system respectively, $L_1$, $L_2$, $L_3$ represent the length of Link1, Link2 and Link3 of the system respectively, $J_1$, $J_2$, $J_3$ represent the inertia of Link1, Link2 and Link3 of the system respectively and g represent gravitational force.

### 4.3 Controllers Tuning Parameters

Table 3 presents the tuning parameters which were obtained through try and error approach,

Table 3: PID, PD and FLC Tuning Parameters

| Variables | Controllers | | |
|---|---|---|---|
| | PID | PD | FLC |
| $K_{p_1}$ | 90 | 76.599 | 20 |
| $K_{p_2}$ | 100 | 205 | 67 |
| $K_{p_3}$ | 75 | 60.795 | 65 |
| $K_{d_1}$ | 90 | 21.999 | 9.275 |
| $K_{d_2}$ | 15 | 13.799 | 13.275 |
| $K_{d_3}$ | 15 | 8.549 | 11.975 |
| $K_{i_1}$ | 1 | - | - |
| $K_{i_2}$ | 1 | - | - |
| $K_{i_3}$ | 1 | - | - |

The Tuning parameters were presented in Table 3, where $K_{p_1}, K_{p_2}, K_{p_3}$ represent proportional gains, $K_{d_1}, K_{d_2}, K_{d_3}$ represent derivative gains and $K_{i_1}, K_{i_2}, K_{i_3}$ represent integral gains, for the controllers.

The Fuzzy logic controller simulation diagrams is presented in figure (7) where the controller is applied to each link of the robot manipulator

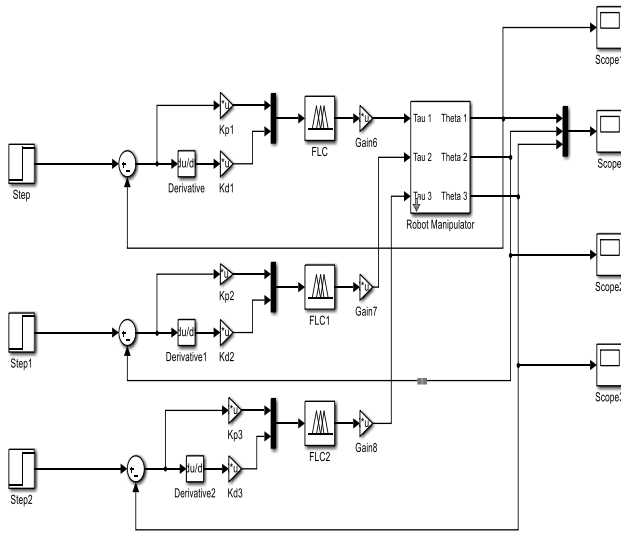

Figure 7: FLC Matlab Simulink Blocks

## 5. OUTPUT RESPONSES

PID, PD and FLC controllers were designed and implemented in Matlab Simulink Environment. In this work the performance of each link were compared with the three controllers.

Figure (8) indicate the responses of the first link of FLC, PD & PID, Figure (9) indicate the responses of the second link of FLC, PD & PID and Figure (10) indicate the responses of the third link of FLC, PD & PID

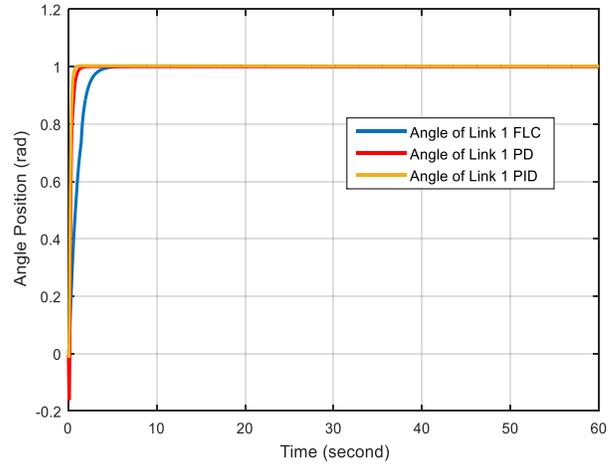

Figure 8: Link 1 Controllers Output Response

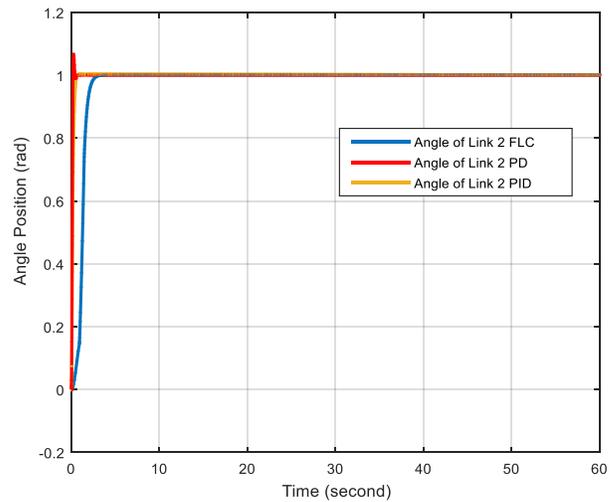

Figure 9: Link 2 Controllers Output Response

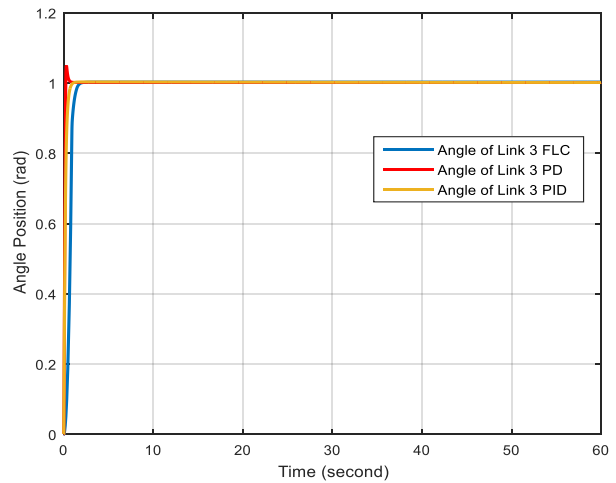

Figure 10: Link 3 Controllers Output Response

## 6. RESULTS COMPARISON

Step input signals with zero step time were used as set-point. The simulation results in figure (8), (9) and (10) show that the output response performance of link1 PID, PD and FLC, link 2PID, PD and FLC and link 3 PID, PD and FLC respectively. The response indicated that all the controllers converged to the set-point

The response parameters, including Rise Time (RT), Settling Time (ST), Overshoot (OS) and Steady State Error (SS) are obtained. Table (4) indicates the comparison of first link of FLC, PD & PID, Table (5) indicates the comparison of second link of the FLC, PD & PID and Table (6) indicates the comparison of third link of the FLC, PD & PID

Table 4: Link 1 Comparison Between PID, PD and FLC

| System Output Characteristics | Controllers | | |
|---|---|---|---|
| | Link 1 PID | Link 1 PD | Link 1 FLC |
| Rise Time (s) | 0.3149 | 0.4513 | 1.8507 |
| Settling Time (s) | 0.6361 | 1.0213 | 3.3064 |
| Overshoot (%) | 0.1395 | 0 | 2.2594e-04 |
| Undershoot (%) | 1.3941 | 16.1508 | 0 |
| Steady State Error | -0.002 | 0 | -0.001 |

In link 1, it was discovered that, the PID (RT=0.3149) (ST=0.6361), and PD (RT=0.4513) (ST=1.0213) outperform FLC (RT=1.8507) (ST=3.3064), the PD (OS=0) (SS=0), and FLC (OS=**2.2594e-04**) (SS=-0.001) outperform PID (OS=0.1395) (SS=-0.002).

Table (5) Link 2 Comparison Between PID, PD and FLC

| System Output Characteristics | Controllers | | |
|---|---|---|---|
| | Link 2 PID | Link 2 PD | Link 2 FLC |
| Rise Time (s) | 0.3011 | 0.1215 | 1.1373 |
| Settling Time (s) | 0.5264 | 1.3642 | 2.4064 |
| Overshoot (%) | 0.1790 | 6.5930 | 7.2402e-04 |
| Undershoot (%) | 0.0028 | 0 | 0 |
| Steady State Error | -0.002 | -0.001 | -0.002 |

In link 2, it was discovered that, the PID (RT=0.3011) (ST=0.5264), and PD (RT=0.1215) (ST=0.3642) outperform FLC (RT=1.1373) (ST=2.4064), the PID (OS=0.1790) and FLC (OS=**7**.2402e-04) outperform PD (OS=6.5930) while the Steady State Error are approximately the same (SS≈-0.002).

Table (6) Link 3 Comparison Between PID, PD and FLC

| System Output Characteristics | Controllers | | |
|---|---|---|---|
| | Link 3 PID | Link 3 PD | Link 3 FLC |
| Rise Time (s) | 0.3583 | 0.1612 | 0.6444 |
| Settling Time (s) | 0.6832 | 0.4484 | 1.4413 |
| Overshoot (%) | 0.1847 | 4.7933 | 1.2071e-04 |
| Undershoot (%) | 0 | 0 | 0 |
| Steady State Error | -0.002 | -0.002 | -0.003 |

In link 3, it was also discovered that, the PID (RT=0.3583) (ST=0.6832) and PD (RT=0.1612) (ST=0.4484) outperform FLC (RT=0.6444) (ST=1.4413), the PID (OS=0.1847) and FLC (OS=**1.2071e-04**) has less overshoot as compared to PD (4.7933) and also the Steady State Error are approximately the same.

In general, it was observed that all the three controllers were able to effectively track the set-point. The FLC has better performance in terms of overshoot as compared to the convectional PD and PID. In terms of Rise Time and Settling Time, however, the PD and PID showed improved performance when compared with the FLC. At lower order system both the convectional PID and FLC can work effectively.

## 7. CONCLUSION

In this study, three different position control methodologies have been designed and analysed on a 3-DOF robot manipulator. The PID, PD and FLC controllers were applied to each link of the robot manipulator and performance comparisons were made using transient and steady state characteristics. The results showed that all three controllers were able to track the setpoint with negligible steady state error. The PID and PD controllers gave better performance in terms of the rise time and settling time while the FLC resulted in decreased overshoot.